# Accelerating Large Kernel Convolutions with Nested Winograd Transformation

Jingbo Jiang[1], Xizi Chen[2*] and Chi-Ying Tsui[1]
Department of Electronic and Computer Engineering, Hong Kong University of Science and Technology, Hong Kong[1]
College of Informatics, Huazhong Agricultural University[2]
Email: jjiangan@connect.ust.hk, xchenbn@mail.hzau.edu.cn, eetsui@ust.hk

*Abstract*— **Recent literature has shown that convolutional neural networks (CNNs) with large kernels outperform vision transformers (ViTs) and CNNs with stacked small kernels in many computer vision tasks, such as object detection and image restoration. The Winograd transformation helps reduce the number of repetitive multiplications in convolution and is widely supported by many commercial AI processors. Researchers have proposed accelerating large kernel convolutions by linearly decomposing them into many small kernel convolutions and then sequentially accelerating each small kernel convolution with the Winograd algorithm. This work proposes a nested Winograd algorithm that iteratively decomposes a large kernel convolution into small kernel convolutions and proves it to be more effective than the linear decomposition Winograd transformation algorithm. Experiments show that compared to the linear decomposition Winograd algorithm, the proposed algorithm reduces the total number of multiplications by 1.4 to 10.5 times for computing 4×4 to 31×31 convolutions.**

## I. Introduction

Deep neural networks have been widely applied to various computer vision tasks over the past decade. While popular neural networks use vision transformers or convolution with stacked small kernels (e.g., 3×3) as their backbone, recent literature [1]–[4] has found that by incorporating training techniques such as re-parameterization, convolution with large kernels such as 31×31 achieves comparable or superior results to state-of-the-art ViTs in many typical downstream computer vision tasks, with lower computational complexity. Literature has also found that large kernel convolution captures more detail than stacked small kernel convolution in applications requiring per-pixel prediction, such as semantic segmentation [5] and image super-resolution [6]. As a result, there has been increasing interest in applying large kernel convolution to different scenarios [7], [8].

Convolution is typically compute-bound [9] because it requires sliding through the entire input feature map to perform repetitive multiplication-and-accumulation (MAC) operations. The Winograd transformation [10] is commonly used to reduce the computational complexity of convolution by replacing some of the expensive multiplications with cheaper operations such as additions or constant multiplications to improve computational throughput. The Winograd algorithm reduces the number of multiplications by using larger transformation matrices for the input feature map, kernel, and output feature map. However, it also introduces numerical instability and computational overhead during data transformation between spatial and Winograd domains [11]. As a result, the common practice for modern ASIC AI processors is to implement a fixed set of Winograd transformation matrices on-chip and only accelerate fixed small kernel (e.g., 3×3) convolution.

Yang et al. [12] and Huang et al. [13] have proposed different accelerator architectures to compute convolution with arbitrary size and stride based on a single set of fixed Winograd transformation matrices. For large kernel convolution operations, they linearly decompose them into a sequence of 3×3 convolution operations by slicing the large input feature maps and kernels into multiple small input tiles and 3×3 kernel tiles, respectively, and then performing 3×3 Winograd convolutions on each pair of input and kernel tiles. However, it has been found that this linear decomposition method does not fully utilize the Winograd transformation to reduce repetitive multiplications between different input tiles, thus not yielding the best multiplication efficiency.

In this work, we propose a *nested Winograd transformation algorithm* that iteratively decomposes a large convolution into a sequence of small (e.g., 3×3) convolutions and prove that it reduces more multiplications than the linear decomposition Winograd algorithm. Intuitively, this is achieved because one small Winograd transformation reduces a certain number of multiplications, and nested Winograd applies fixed small Winograd transformations more times to the data than the linear decomposition Winograd transformation. To demonstrate the effectiveness of the proposed algorithm, we propose an accelerator architecture and runtime for utilizing nested Winograd transformation to accelerate large kernel convolution and verify its effectiveness using FPGA. We observe that nested Winograd reduces the number of multiplications by 1.4 to 10.5 times compared to the linear decomposition Winograd algorithm when running convolution with kernel sizes ranging from 4×4 to 31×31. We also show that compared to a previous linear decomposition Winograd accelerator running FSRCNN-s for image super-resolution [14], our proposed nested Winograd accelerator achieves an overall 1.27 times throughput improvement. In summary, the contributions of this paper are as follows:

- A nested Winograd transformation algorithm is proposed to accelerate the execution of convolution with large kernel sizes and proved to outperform the state-of-the-art linear decomposition Winograd algorithm.
- An accelerator architecture and runtime are proposed to accelerate native convolution with arbitrary kernel sizes using nested Winograd transformation.

## II. Background

### A. Winograd Algorithm

A 2D stride-1 native convolution correlates $M$ channels of input feature maps $x$ of size $H \times W$ with $N$ groups of $M$-

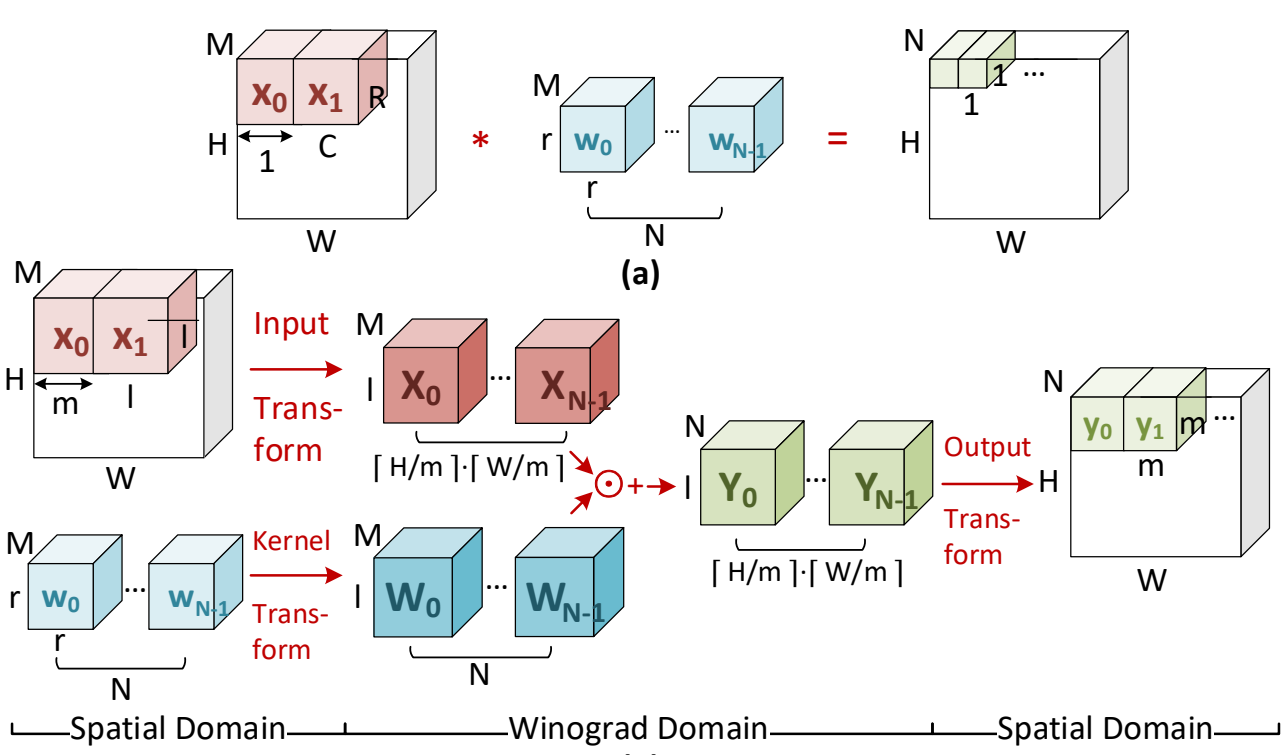

Fig. 1. The computation flow of (a) the native convolution and (b) the Winograd convolution $F(m \times m, r \times r)$.

channel kernels $w$ of size $R \times C$ to produce $N$ channels of output feature maps $y$. For each channel of the $R \times C$ kernel, ```the native convolution takes an $R \times C$ tile from the input feature map, performs a MAC operation on it, and produces one output pixel. The input feature map window then slides by 1 to take the next input tile.

In contrast, Winograd convolution takes a larger $l \times l$ input tile $x$ from the input feature map, transforms it along with the kernel into the Winograd domain to produce two $l \times l$ tiles, and then performs element-wise matrix multiplication (EWMM) between these two $l \times l$ tiles to create an $l \times l$ output tile. Finally, the output tile is transformed back into the spatial domain to produce an $m \times m$ output tile. Stride $m$ is used to sample the next input tile from the input feature map. This procedure is denoted as $F(m \times m, r \times r)$ [10], where $r \times r$ is the kernel size of the native convolution (in this example $R = C = r$) and $l = m + r - 1$. This process is illustrated in Fig. 1. Let $B, G$ and $A$ denote the input, kernel, and output transformation matrices, respectively, and let $\odot$ represent the EWMM. The 2D Winograd convolution can be formulated as

$$y = A^T (B^T x B \odot G w G^T) A \tag{2.1}$$

For 1D Winograd convolution applied to the input vector $x$ and kernel vector $w$, (2.1) reduces to

$$y = A^T (B^T x \odot \mathrm{G}w) \tag{2.2}$$

and is represented as $F(m, r)$. By comparing (2.1) and (2.2), it can be seen that multi-dimensional Winograd convolution is constructed by applying the 1D Winograd transformation to each axis of the input tensor, kernel, and output tensor, respectively. Therefore, $F(m, r)$ is referred to as the *Winograd base*, reflecting its role as the base operator for constructing the multi-dimensional Winograd algorithm. The transformation matrices of a commonly used Winograd base $F(2,3)$ is

$$B^T = \begin{bmatrix} 1 & 0 & -1 & 0 \\ 0 & 1 & 1 & 0 \\ 0 & -1 & 1 & 0 \\ 0 & 1 & 0 & -1 \end{bmatrix} G = \begin{bmatrix} 1 & 0 & 0 \\ 1/2 & 1/2 & 1/2 \\ 1/2 & -1/2 & 1/2 \\ 0 & 0 & 1 \end{bmatrix}$$

$$A^T = \begin{bmatrix} 1 & 1 & 1 & 0 \\ 0 & 1 & -1 & -1 \end{bmatrix} \tag{2.3}$$

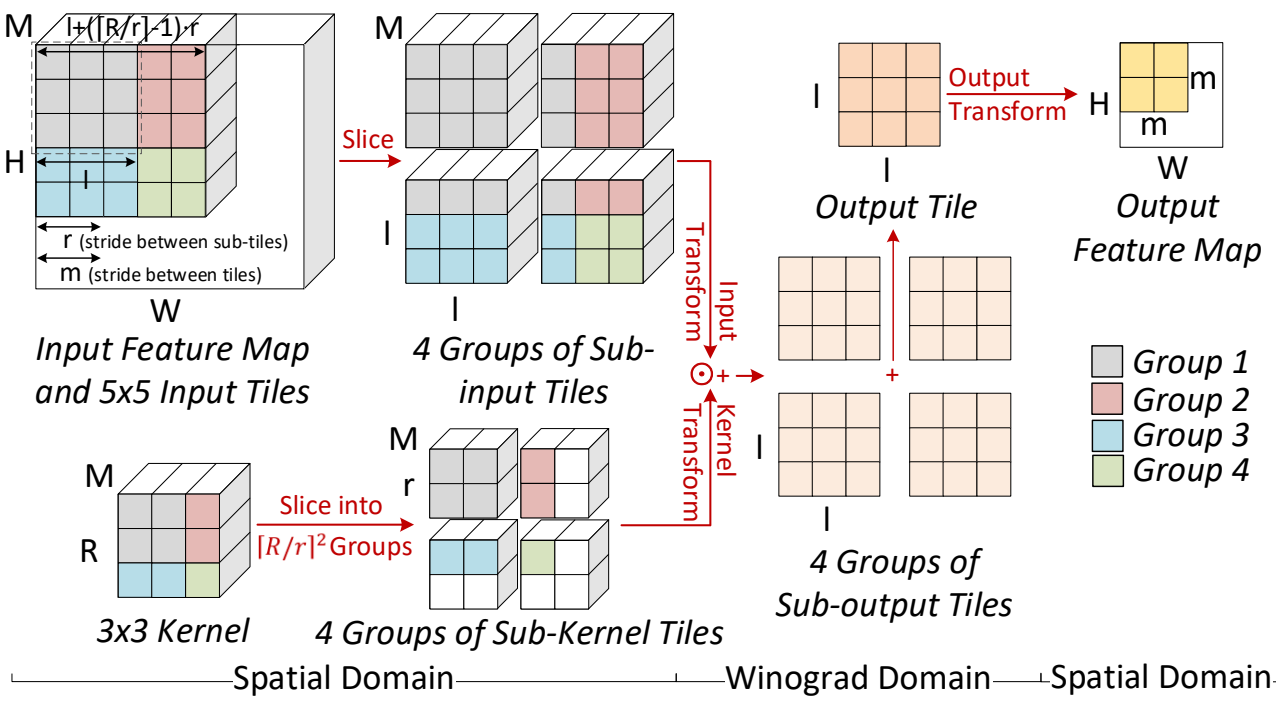

Fig. 2. Accelerating the convolution of a 5×5 input tile with a 3×3 kernel using linear-decomposition Winograd $F(2 \times 2, 2 \times 2)$.

The Winograd algorithm reduces the number of multiplications by exploiting data overlaps between adjacent input tiles in native convolution [15], which would otherwise require computing repetitive multiplications on the same data. The Winograd algorithm transforms the MAC operation from the real number field to a finite field, allowing it to explore the data overlaps and use additions and constant multiplications to replace some of the repetitive multiplications. If there are no data overlaps between adjacent input tiles, the Winograd algorithm reduces to $F(1, r)$ and converges to native convolution.

## B. *Linear-decomposition Winograd algorithm*

Accelerating convolutions with different kernel size $r \times r$ requires the use of different Winograd base. However, instantiating multiple Winograd bases on-chip reduces area efficiency. Linear-decomposition Winograd algorithm provides a solution on accelerating convolution with $R > r$ on a fixed $F(m, r)$.

Fig. 2 illustrates the procedure for accelerating a 3×3 convolution using the linear-decomposition Winograd algorithm with $F(m = 2, r = 2)$. The $R \times R$ kernel is divided into $\lceil R/r \rceil^2 = 4$ groups of non-overlapping sub-kernel tiles, each with a size of $r \times r$ and can be accelerated with $F(m = 2, r = 2)$. The input tile, sized $h \times w$ where $h = w = l + (\lceil R/r \rceil - 1) \cdot r = 5$, is also divided into the same number of sub-input tiles. Unlike sub-kernel tiles, adjacent sub-input tiles have a size of $l \times l$ and are sliced using a stride-$r$ sliding window, creating data overlap between them. EWMM is then performed between these sub-input tiles and sub-kernel tiles to produce $\lceil R/r \rceil^2$ groups of sub-output tiles. These sub-output tiles are further summed together to produce an $l \times l$ output tile. Note that in this step, $M$ input channels are accumulated into one output channel. Finally, the $l \times l$ output tile is transformed back into the spatial domain to produce an $m \times m$ output tile. The sliding window on the input feature map slides by $m$ to generate the next input tile.

Although the Winograd algorithm is applied to each pair of sub-input and sub-kernel tiles to reduce the number of multiplications, the linear-decomposition algorithm does not explore the data overlap between sub-input tiles. Methods exist to apply additional Winograd transformations on these overlapped data to further reduce the number of repetitive multiplications.

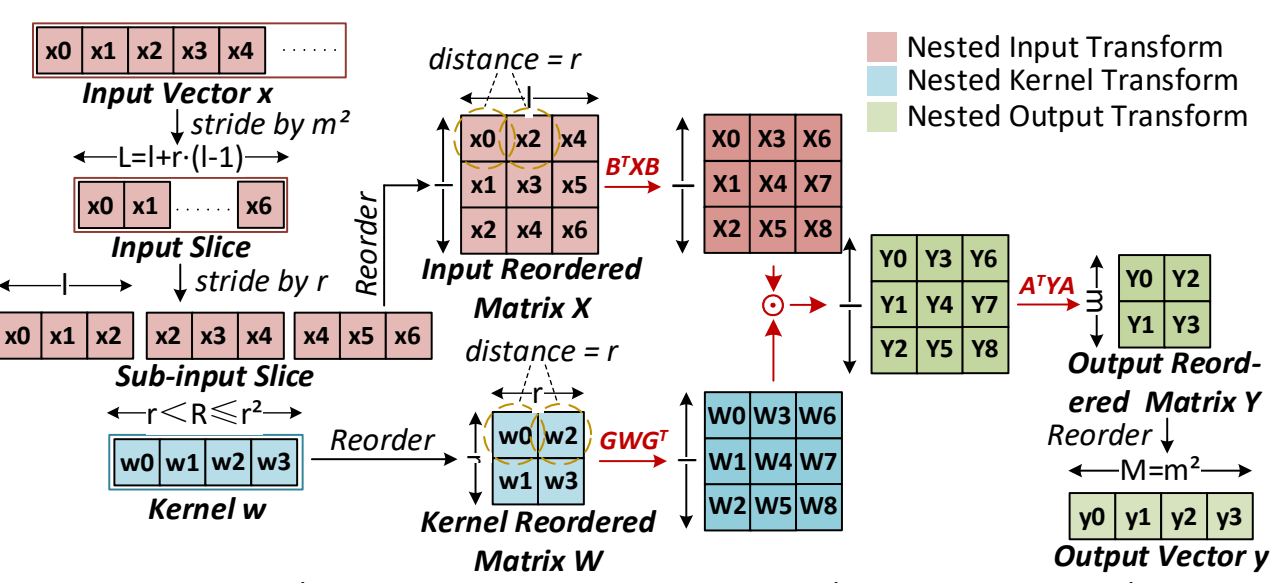

Fig. 3. Example of accelerating 1D convolution with a length-4 kernel using nested Winograd algorithm based on $F(2,2)$.

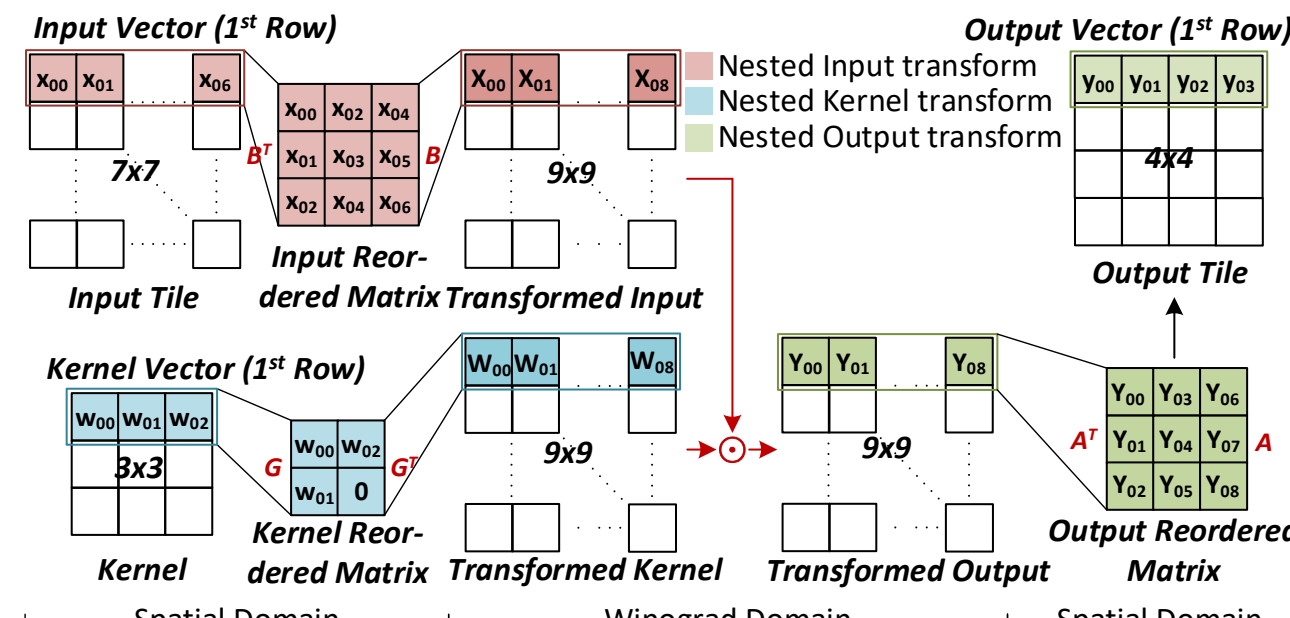

Fig. 4. Accelerating a 2D convolution with a 3×3 kernel using nested Winograd algorithm based on $F(2,2)$.

## III. The Nested Winograd Algorithm

### A. *The nested decomposition algorithm*

*Nested decomposition* is a general method that uses high-dimensional fast algorithm transformations, such as 2D Winograd or 2D fast Fourier transformation (FFT), to accelerate low-dimensional convolution (e.g., 1D). This is achieved by re-expressing the input vector $x$ with length $N = N_1 \cdot N_2$ into a matrix $X$ of size $N_1 \times N_2$. Then, a 1D fast algorithm transformation is applied to each column follows by each row of $X$. This procedure can be formulated as

$$(A \cdot X) \cdot A^T \tag{3.1}$$

where $A$ represents the transformation matrix of the 1D fast algorithm. Note that (3.1) has a similar form to the 2D Winograd transformation shown in (2.1) but is used to accelerate 1D convolution. Different data pre-processing procedures, such as padding and repeating, must be applied to $X$ to align it with the corresponding re-expressed kernels for various types of long convolution problems.

Nested decomposition has been applied to various types of convolution problems throughout history. For example, in 1964, Cooley and Tukey applied nested decomposition to accelerate discrete Fourier transform (DFT) [16]. Later, in 1977, Agarwal and Cooley applied nested decomposition with Winograd algorithm to the cyclic convolution [17]. These prior works inspired us to develop a method for applying nested decomposition with Winograd algorithm to the convolution in machine learning.

### B. *The proposed nested Winograd algorithm*

We begin by using an example to illustrate how to apply our proposed nested Winograd transformation to 1D convolution and then extend it to 2D convolution. Figure 3 illustrates the process of using nested Winograd with $F(m = 2, r = 2)$ to accelerate a 1D convolution with kernels of length $R = 4$.

The first step is to perform the *nested input transformation*. This involves applying a stride-$m^2$ sliding window to slice the long input vector into length-$L$ input slices, where $L = l + r \cdot (l - 1)$ and $l = m + r - 1$. Then, a stride-$r$ sliding window is applied to further slice the input slices into length-$l$ sub-input slices, which are arranged column-by-column to form the *input reordered matrix* $X$ of size $l \times l$. The 2D Winograd input transformation $B^T \cdot () \cdot B$ is then applied to $X$ to transform it into the Winograd domain.

The second step is to perform the *nested kernel transformation*. This is achieved by reshaping the length-$R$ kernel into a $r \times r$ sized *kernel reorder matrix* $W$ and using the 2D Winograd kernel transformation $G \cdot () \cdot G^T$ to transform $W$ into the Winograd domain.

The nested Winograd algorithm requires data alignment between the input data and the kernel. This means that adjacent columns in the input reordered matrix must have the same *index distance* as the kernel matrix. For instance, as illustrated by the dashed yellow circle in Fig.3, the horizontal elements in the kernel reordered matrix have index distance $r = 2$. This necessitates that the sliding window applied to the input slice has a stride equal to $r$. Given that $F(2,2)$ requires the sub-input slice to have a length of $l$, the window length of the input slice is calculated as $L = r \cdot (l - 1) + l$. Finally, the input vector $x$ should stride by $m^2$ because nested Winograd produces $m^2$ output data elements when convolving kernel with one input slice.

The final step involves performing an EWMM between the input and kernel reordered matrices. This is followed by a *nested output transformation*, where a 2D Winograd output transformation $A^T \cdot () \cdot A$ is applied to produce the output reordered matrix $Y$. Finally, the output reordered matrix is flattened into a length-$m^2$ output vector.

Winograd algorithm specifies that the kernel transformed by $F(m, r)$ must have a length no greater than r. Similarly, the kernel transformed by $F(m \times m, r \times r)$ used in nested Winograd must have a length $R$ no greater than $r^2$. To accelerate convolution with a kernel length greater than $r^2$, the kernel should first be reshaped into a tensor and then be transformed with a multi-dimensional Winograd algorithm [18]. If $R$ is not equal to an integer power of $r$ (i.e., $r, r^2, r^3, \ldots$), zero-padding should be performed at the end of kernel $x$. In general, if an n-dimensional Winograd transformation is used, the nested Winograd algorithm can accelerate convolution with a kernel length no greater than $r^n$, consume $l^n$ multiplications and produce $m^n$ output data elements for computing one input slice.

The above description outlines the general procedure for aligning data between the input and kernel reordered matrices for all $F(m, r)$. However, it is recommended to use $F(m, r)$ with $m = r$ when possible. Otherwise, the output vector $y$ may include redundant data elements that should be discarded, potentially reducing the computational efficiency.

The procedure for accelerating 2D native convolution using the nested Winograd algorithm is illustrated in Fig.4. First, the previously described nested input and kernel transformations are applied to all rows follows by columns of the input tiles and kernels, respectively. Then, EWMM is performed between the transformed input and kernel, and produce the transformed output matrix. Finally, the first row of the transformed output matrix is transformed back to the space domain using the nested output transformation. The result is then reshaped to form the output vector. This process is iteratively performed on all rows followed by all columns of the transformed output matrix to produce the final output tile.

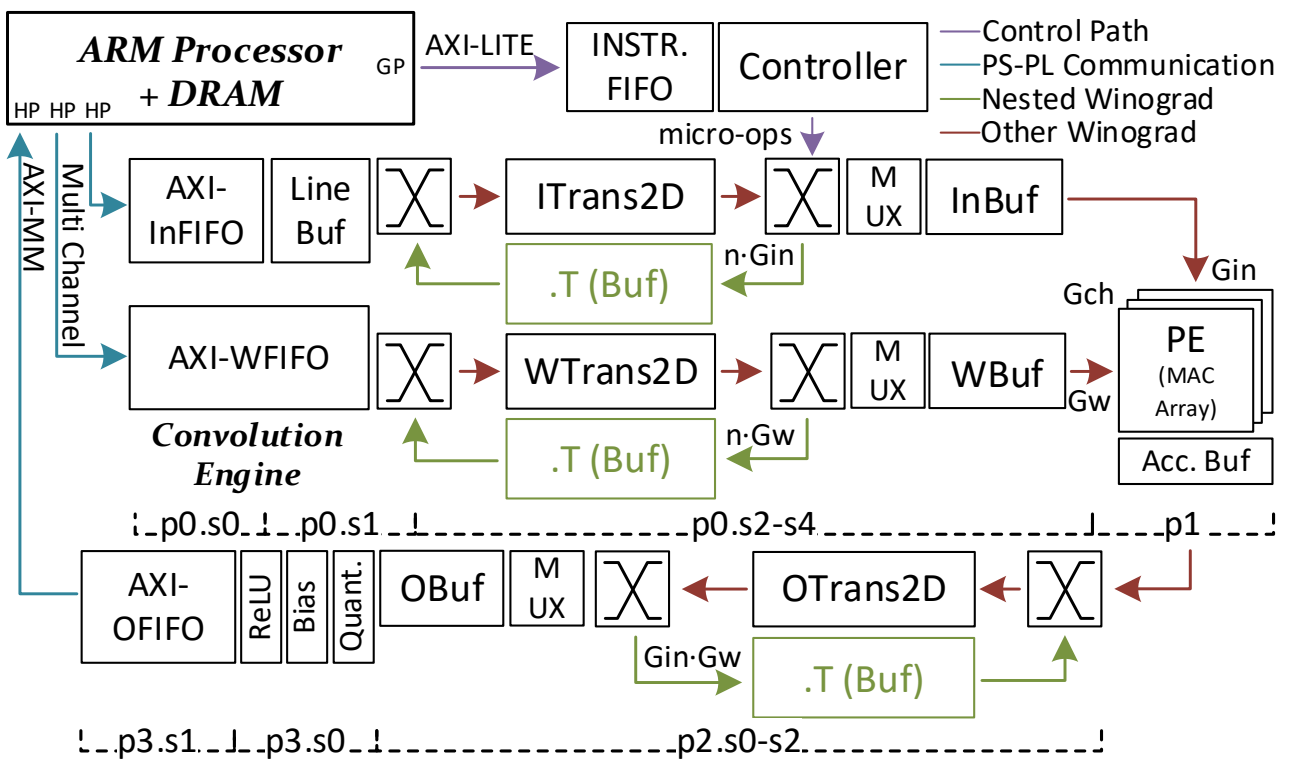

Fig. 5. The architecture of the accelerator and the convolution engine

### C. *Multiplication Complexity Analysis*

We provide an algorithm analysis to demonstrate the advantages of using nested Winograd. The multiplication complexity is defined as the number of multiplications required to produce one output data element in a convolution. Similar to the multiplication complexity analysis of the vanilla Winograd algorithm [15], the number of additions and constant multiplications incurred by the Winograd transformation are considered as overhead and are not included in this analysis.

To derive the multiplication complexity, considering a convolution of a length-$R$ kernel with an infinite length input vector using $F(m,r)$. The linear decomposition Winograd algorithm slices the length-$R$ kernel into $R/r$ sub-vectors and applies $F(m,r)$ to them individually to produce a length-$m$ output kernel. Since each $F(m,r)$ transformation requires $l = m + r - 1$ multiplications, the overall multiplication complexity is

$$O((l/m)\cdot(R/r)) \tag{3.2}$$

For nested Winograd, we only analyze Winograd base with $m = r$ to simplify the discussion, the $m \neq r$ case can be derived in a similar manner. Assuming the reordered tensor of nested Winograd has $b = log_r R$ dimensions. Since applying $F(m,r)$ to one length-$r$ kernel consumes $l$ multiplications and produce $m$ output elements, the overall algorithm complexity for nested Winograd is the product of the multiplication complexity in all dimensions, given by

$$O\left((l/m)^{\log_r R}\right) \tag{3.3}$$

It is hard to directly compare the complexity of (3.3) and (3.2) at this stage. Thus, we further approximate (3.3) with the order-1 Taylor expansion at point $r$ and we get

$$O\left(\frac{l}{m\cdot r}\cdot\log_r(\frac{l}{m})\cdot R+\alpha\right) \tag{3.4}$$

where $\alpha$ represents the constant term. If we apply $l = m + r - 1$, then divide (3.4) by (3.2) and let the result smaller or equal to 1, we get $m + r - 1 \leq m \cdot r$, which is valid for $m$ and $r$ belongs to any positive integer. It means that nested Winograd almost always has a gentler slope than the linear decomposition Winograd algorithm. However, as (3.4) also has a constant term which may let nested Winograd be less efficient than OLA-Winograd, a simulation on some common cases is performed and the performance difference between these two algorithms is summarized in Section V.

**Algorithm 1**: Decomposition Algorithm

**Input:** $F(M,R)$ to be decomposed
**Output**†**:** $ET$ containing $\{F(m,r),\otimes\}$, $M$
*Initialize ET and M*
1: $ET = [F(M,R),]$
2: $M = m$
*Procedure to decompose* $F(M,R)$*:*
3: **for** ($F$ in $ET$) and ($F$ is a Winograd base) **do**
4: **while** $F$ has $R > r$ **do**
6: $F_0, F_1 =$ NestedWinogradDecompose($F$)
7: Replace $F$ in $ET$ with $[F_0, F_1, \otimes]$
8: Let $M = M^2$
10: **end while**
11: **end for**
13: **function** NestedWinogradDecompose ($F(M,R)$)
14: n=0
15: **while** $R < r^{n+1}$ **do** n=n+1; **end for**
16: **return** $F_0(M, r^n), F_1(M, r^n)$
17: **end function**
18: **return** $ET, M$

† $ET$ denotes expression tree, $M$ denotes output length.

## IV. The Accelerator Design

### A. *Overview of the Accelerator Design*

We designed an accelerator architecture and implemented it with Xilinx ZYNQ FPGA to evaluate the effectiveness of the nested Winograd algorithm. The accelerator is composed of two parts: an on-chip Arm processor that implements a runtime for decomposing a large kernel convolution into multiple fixed $F(m,r)$ Winograd convolutions, and a fabric logic array that implements a convolution engine for executing fixed $F(m,r)$ Winograd convolutions.

The accelerator is programmed by calling self-defined C++ functions such as *Conv2d()*, in the Xilinx Software Development Kit. When the *Conv2d()* function is called, the Arm processor decomposes it into a sequence of *F(m,r)* Winograd convolution instructions and sends them to the convolution engine via the AXI-LITE bus through the Xilinx General Purpose (GP) port. Upon receiving an instruction, the convolution engine decodes it into multiple micro-operations which are then dispatches to different compute pipelines for execution. Once input data is streamed in and all executions are completed, the convolution engine streams the output data from its on-chip buffer to the external DRAM via the AXI-MM bus through Xilinx High Performance (HP) port. The Winograd convolution instruction is then retired and this process is repeated until all layers of a CNN have been processed.

### B. *The Convolution Engine*

The micro-architecture of the convolution engine is illustrated in Fig.5, which majorly contains the following modules: an instruction FIFO buffering the instruction sent from the Arm processor; a Controller decoding the instruction into micro-operations and dispatching them to the compute pipelines; four multi-stage (s0-s4) compute pipelines including the forward transformation pipeline (p0), the general matrix multiplication (GEMM) pipeline (p1), the backward transformation pipeline (p2), and the post-processing pipeline (p3); scratchpad memories (SPMs) including InBuf, WBuf, Acc.Buf, OBuf isolating the compute pipelines; AXI-InFIFO, AXI-WFIFO and AXI-OFIFO exchanging data between the Arm processor and the convolution engine; a Line Buffer unrolling input tensor into overlapped input tiles; pipelined Input, kernel and output 2D Winograd transformation units implementing fixed Winograd base; matrix transpose blocks (.T) implemented as a shift-register based buffer to perform Winograd transformations in a loop for accelerating 2D convolution with nested Winograd; a processing element (PE) containing $G_{ch}$ channels of $G_w \times G_{in}$ number of MACs connected with a 2D broadcast network to perform output stationary GEMM; fixed-functional units including Relu, Bias, and Quantization for post-processing.

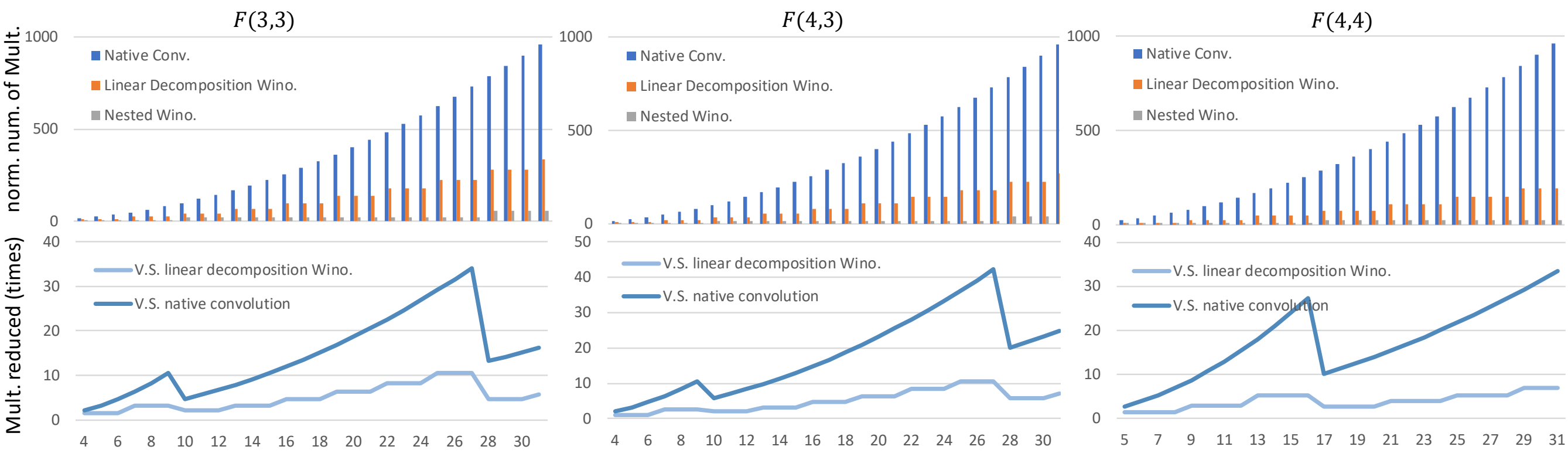

Fig. 7. Multiplication complexity comparison between the native convolution, linear decomposition Winograd and nested Winograd algorithm.

The total number of execution cycles required to complete one instruction varies among the four compute pipelines (p0-p3). This variation can be attributed to the fact that the nested Winograd transformation in p0 can require multiple iterations of $F(m \times m, r \times r)$. During each iteration, data is transposed by the .T block and returned to the Trans2D block until completion. p1 consumes the transformation results while p0 produces and writes them to InBuf and WBuf, and OBuf. The production and consumption throughput does not need be equal. p1 is often the through bottleneck of all compute pipelines due to the limited number of power-hungry MACs imposed by the thermal design power (TDP) requirement. p1 is often the throughput bottleneck of all compute pipelines because the thermal design power (TDP) requirement of the accelerator often limits the number of power-hungry MACs. Multiple Winograd convolution instructions can be pipelined naturally to maximize the resource utilization rate of p0 to p3.

### C. The Decomposition Algorithm

The decomposition algorithm running in runtime on the Arm processor decomposes a *Conv2d()* function that executes a convolution with arbitrary kernel size into a stream of *F(m,r)* Winograd convolution instructions. Alg. 1 shows the procedure of this algorithm, which will be illustrated with an example of decomposing 5×5 convolution into a stream of $F(m = 2, r = 2)$ Winograd convolution instructions.

The process of using the Winograd algorithm to accelerate 5×5 native convolution can be represented as $F(M, R = 5)$, where $M$ denotes the overall output length that needs to be calculated using Alg. 1. The decomposition algorithm iteratively checks if the given $F(M, R)$ has $R > r$. If this condition is met, it decomposes $F(M, R)$ using nested Winograd algorithm:

$$F(M, 5) = F(M', 2) \otimes F(M', 2) \qquad (4.1)$$

where $\otimes$ represents the procedure of nesting decomposition. (4.1) formulates the nested Winograd algorithm, indicating that an $F(M, 5)$ Winograd convolution can be achieved by applying a nested decomposition between two $F(M', 2)$ Winograd convolutions. In the runtime C++ program, $F(M', 2) \otimes F(M', 2)$ is represented as an *expression tree* implemented as a linked list $[F(M', 2), F(M', 2), \otimes]$. Once the decomposition procedure is complete, the linked list is read out in *reverse polish order* to send the Winograd convolution instructions. The next step is to determine the value of $M$. Since $F(M', 2)$ cannot be further decomposed, $M'$ is set to $m = 2$. Then, we compute backward, squaring $m$ for each iteration until reaching $F(M, R)$, and finally obtaining $M = m^2 = 4$.

## V. Experiments

### A. Multiplication Complexity Simulation

To compare the multiplication complexity between nested Winograd and linear decomposition Winograd algorithm, we simulated 2D convolutions with different kernel sizes ranging from 3×3 to 31×31 with $F(3,3)$, $F(4,3)$ and $F(4,4)$. Winograd base larger than $F(4,3)$ is rarely used due to the increased transformation overhead [11].

The simulation results summarized in Fig.7 demonstrate that the nested Winograd algorithm reduces 1.17 to 10.56 times more multiplications than the linear decomposition Winograd algorithm and reduces 2.07 to 44.25 times more multiplications than the native convolution. The gap increases as the kernel size becomes larger, and the zig-zag shaped multiplication reduction curve results from the fact that nested Winograd based on $F(m, r)$ requires padding the kernel height and width to the power of $r$.

TABLE I. GOPS SPEED UP ON DIFFERENT CONVOLUTIONS

| Conv. Type | *From* | *Conv. Shape in (Cout, Cin, H, W)* | *GOPs[b] Nested Win.* | *GOPs L.D.[c] Win.* | *Speed up* |
|---|---|---|---|---|---|
| Conv2D 9×9[a] | SRCNN Layer-1 | (64,1,256,256) | 3503 | 1063 | 3.29 |
| Depth-wise 7×7 | PNasNet | (54,54,83,83) | 540 | 384 | 1.41 |
| Conv2D 5×5[a] | SRCNN Layer-2 | (32,64,256,256) | 991 | 692 | 1.43 |
| Conv2D 5×5[a] | SRCNN Layer-3 | (1,32,256,256) | 186 | 130 | 1.43 |

a. These three layers concludes all the layers of the SRCNN.

b. GOPs = (#multiplication + #addition) / execution time when using native convolution.

c. L.D. is short for linear decomposition.

### B. *End-to-end Performance Evaluation*

We compare the throughput of running nested Winograd and linear decomposition Winograd algorithm on the accelerator designed in Section IV with CNNs that have kernel size ranging from 5×5 to 9×9. The accelerator is implemented on the Xilinx ZCU102 board containing an Arm dual-core A53 processor with a DDR4-2666 providing sufficient bandwidth in off chip data accessing. The convolution engine is implemented to run at 200MHZ which is consistent with [2]. The convolution engine is implemented with $F(3,3)$ and the $G_{ch}$,$G_{in}$,$G_w$ are set to be 25, 6, 6.

We compared the throughput of running the nested Winograd algorithm and the linear decomposition Winograd algorithm on the accelerator designed in Section IV with different CNNs. The accelerator was implemented on a Xilinx ZCU102 board containing an Arm dual-core A53 processor with DDR4-2666, providing sufficient bandwidth for off-chip data access. The convolution engine is implemented to run at 200MHZ with 16-bit data path. The Winograd base is set to $F(3,3)$ and the $G_{ch}$,$G_{in}$,$G_w$ are set to be 25, 6, 6.

The acceleration results of four single convolution layers are demonstrated in Tab. I. The 5×5 and 9×9 convolution layers were chosen from the acceleration result of SRCNN [20], while the 7×7 depthwise convolution layer is chosen from PNasNet [21]. We further compare the end-to-end acceleration results with a previous work [14], which design an FPGA accelerator running FSRCNN-s to upscale an image from 1920×1080 to 3840×2160. They propose an FTConv algorithm to decompose all the layers in FSRCNN-s to 5×5 convolutions, then use linear decomposition Winograd to accelerate it. The decomposed FSRCNN-s has 86.1% MACs coming from the 5×5 convolution. The results summarized in Tab. II show that our accelerator achieves 1.27 times end-to-end speed up compared with [14] with around the same number of DSPs. We did not observe a PSNR drop by adopting nested Winograd in this experiment.

## VI. CONCLUSION

In this work, a nested Winograd algorithm is proposed for accelerating convolution with large kernel size. The multiplication complexity is reduced when comparing with existing linear decomposition Winograd algorithm. Simulation results show 1.4 to 10.5 times speed-up in executing convolution with kernel size from 4×4 to 31×31 using $F(3,3)$ Winograd transformation.

TABLE II. THROUGHPUT COMPARISON ON FSRCNN-S NETWORK

| | #LUT | #DSP | BRAM (Mb) | Frame Rates |
|---|---|---|---|---|
| [14] | 172K(63%) | 746(30%) | 10.9(34%) | 120.4(fps) |
| **Ours** | 184K(67%) | 748(30%) | 9.8(30%) | 153.1(fps) |

## ACKNOWLEDGMENT

We would like to extend our sincere gratitude to the Hong Kong AI Chip Center for Emerging Smart Systems (ACCESS) for their pivotal support to our work.